# SPEEDY OBJECT DETECTION BASED ON SHAPE

Y. Jayanta Singh, Shalu Gupta

Dept. of Computer Science & Engineering and Information Technology,
Don Bosco College of Engineering and Technology of Assam Don Bosco University
jayanta@dbuniversity.ac.in
shalu2324@gmail.com

## ABSTRACT

*This study is a part of design of an audio system for in-house object detection system for visually impaired, low vision personnel by birth or by an accident or due to old age. The input of the system will be scene and output as audio. Alert facility is provided based on severity levels of the objects (snake, broke glass etc) and also during difficulties. The study proposed techniques to provide speedy detection of objects based on shapes and its scale. Features are extraction to have minimum spaces using dynamic scaling. From a scene, clusters of objects are formed based on the scale and shape. Searching is performed among the clusters initially based on the shape, scale, mean cluster value and index of object(s). The minimum operation to detect the possible shape of the object is performed. In case the object does not have a likely matching shape, scale etc, then the several operations required for an object detection will not perform; instead, it will declared as a new object. In such way, this study finds a speedy way of detecting objects.*

## 1. INTRODUCTION

An in-house object detection system for visually impaired, low vision personnel is require to support the user to act independently. The output of such system can be directed to Bluetooth devices in the form of sound. The details description of the detected objects and its severity levels could be provided to alert at any time of difficulties. This will help to deal with objects come as hindrances in front of target user. The study concentrated to provide a minimum storage space and speedy ways of object detections. This in-house auditory object detection system can provide an alert system with severity of dangerous and harmful object ahead. Combinations of techniques of digital image processing and speedy database management system are used. It aims to develop system with minimum infrastructure and cost so that it is feasible to embed into low cost devices. The possible shape of the query image is not matching with shapes of objects in the available database (training), the remaining several operation of object detection: segmentation, cleaning, normalization, detection etc will not perform. The minimum operation to detect the possible shape of the object is performed. This will save computation time. This report contains does not contains the entire result of the study as the study is ongoing one.

## 2. PREVIOUS WORK

Many Electronic Travel Aid (ETAs) devices available for visually impaired people. For example, there are ETA devices in USA, such as LaserCane[1], NavBelt[2], PeopleSensor[3], GuideCane[4], Tyflos[5], Binaural Sonic Aid[6], v0ICe in Netherlands[7], 3-D Space Perceptor in Canada[8], NSOB in Japan[9], ESSVI in Italy[10], Navigation Assistance Visually Impaired(NAVI) in Malaysia[11], AudioMan[12], SoundView[13] and [14] divides the blindness techniques into sensory substitution, sonar based ETAs and camera based ETAs.

Most of the devices works on GPS and MAPs which is not highly possible in localities of middle class families. In most of study, all the pre-processing operations are performed for the object that does not available in the training database (or a new object). This study will filter out the object based on shape and scale and then decided whether to execute the further operation or not.

## 3. SYSTEM DESIGN

Object can know by components [15]. This is the primary way of classifying objects by identifying their components and the relational properties among these components. The other features like texture, size or colour can be use to distinguish the close similar objects wherever necessary. The general approach architecture of the study and a brief description of the components of the architecture are provided in figure1.

### 3.1 Camera with remote control

The objects will be captured by the camera. This camera can be controlled by customized design remote control device which is having port to connect the camera. The remote control device is of palm size only. The remote control will have five facility buttons with defined facilities. Auto detection of an image will be enabling by double clicking the button2. Different forms of Wi-Fi enable camera which has a flexible-removable neck or pen shape wearable camera can be used to capture images. We use the concept of ‘double click’ and ‘single click’ to switch between functionalities of this remote control device. The proposed functionalities of the remote control device are provided. Software interface will be used to mimic for these buttons for the initial phase of the experiment.

Sample of several facility buttons required for operations:

1: Initiate the object detection system
2: Detected objects
3: Revisiting and learning the details of the objects (for beginners)
4: Alert on difficulties
5: Home (return to home)

### 3.2 In-house detection system

The scenes are captured through the camera and processed and detected by using several techniques. The details of the technique used for the study is describe in alter part of this report.

### 3.3 Bluetooth device and the Central Speaker

The audio format of the description of the detected objected will played through this Bluetooth device. The central speaker will be used in emergency, difficulties and testing purpose. In case the proper actions are unable to perform by the user (minor or aolge age), any nearby people can also know the result of the detected object through this central speaker.

### 3.4 Central Wi-fi system

Architecture is setup in the wi-fi environment where the captured scene (images) can be send to the processing system and the result can be send back to Bluetooth devices without wires in automatic ways.

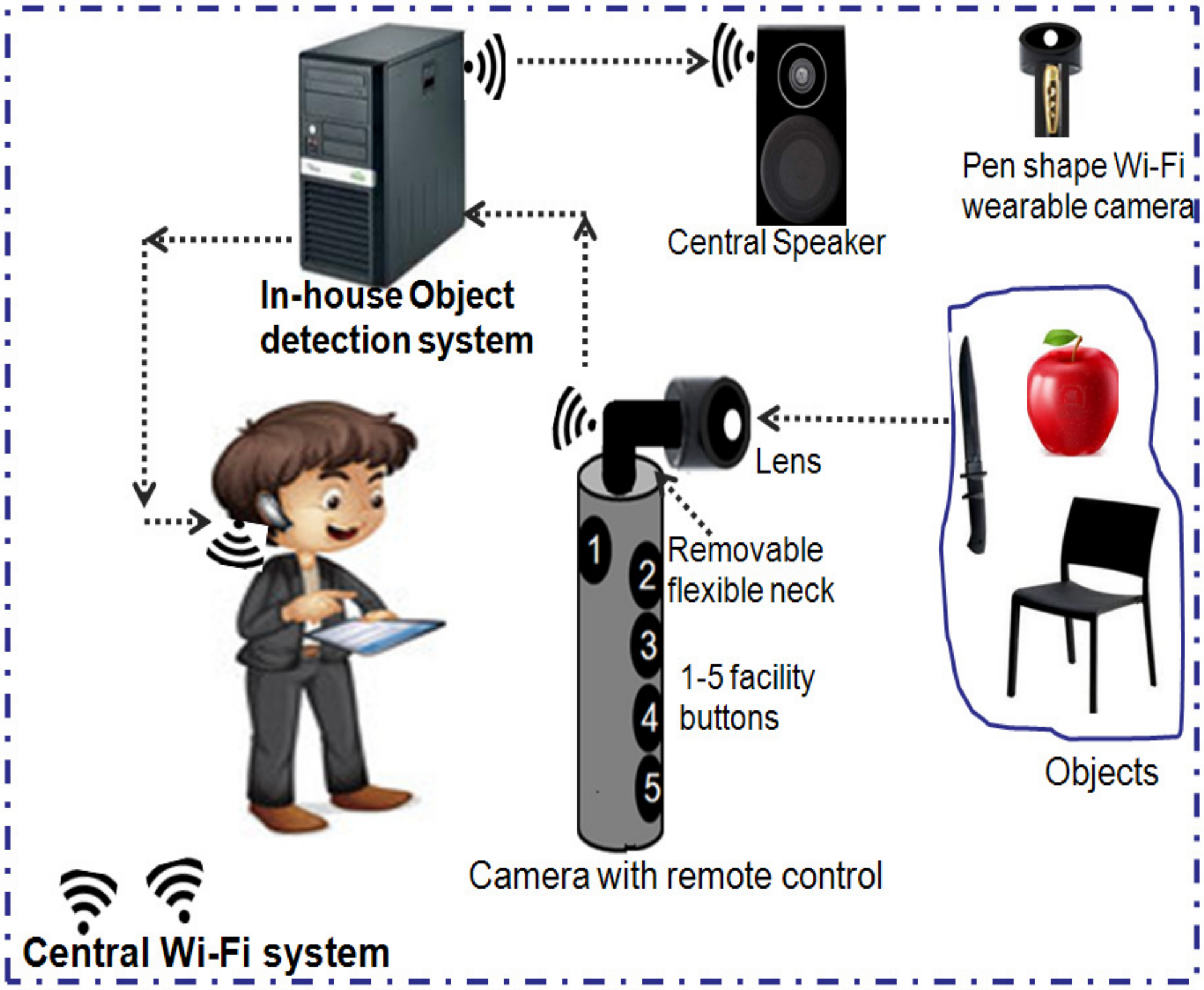

Figure1. Physical layout of an audio system for in-house object detection

## 4. THE EXPERIMENTS

This report is part of development of an audio system for in-house object detection. The study was executed aiming to provide speedy selection of features during the training time and speedy detection of objects during the testing time. This study implemented several different techniques to achieve the aim. Pre-processing of images is done prior to segmentation of objects. Thinning and proper normalization the input images are performed. For this portion of study the binary image are taken input and then transformed into a set of basic digital primitives (lines, arcs, etc) that lie along their axes [16]. Initially we are taking the different shapes sample from the MPEG-7 database [17]. The proposed algorithms used during the training and testing phases of this study are provided below.

### 4.1. Detect object based on the shapes

We take the features of the objects based on the basic geometrical shapes such as line, rectangle, square, circle, scale of the object (discussing in the next point) and other required parameters. Some examples of sample shapes are given in figure2. Lists of clusters of the features of the objects are formed based on shape followed by the scale size of the objects. A global index is created and updates each after a feature is extracted. The index indicates the location of the clusters containing a specific shape and scale. Cluster1 may contain features of all objects having the shape of a rectangle (3), for example, Table, Chair, TV, and Laptop, Mobile etc, shown in figure3. Cluster2 may contain the features of objects having shapes of pointed edges (1) like Pin, pencil, Nail etc. Size of these objects could be different. Few objects could present in both of the cluster1 and 2.

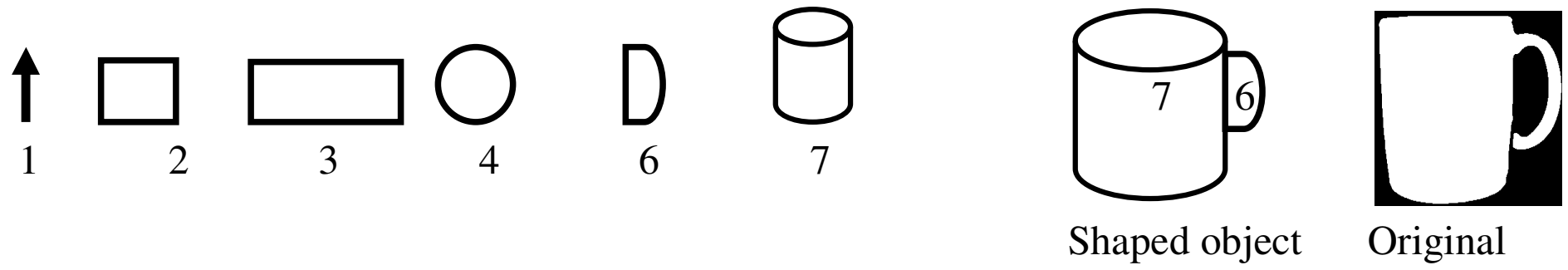

Figure2. Samples of the 7 shapes, a shaped object and an original object

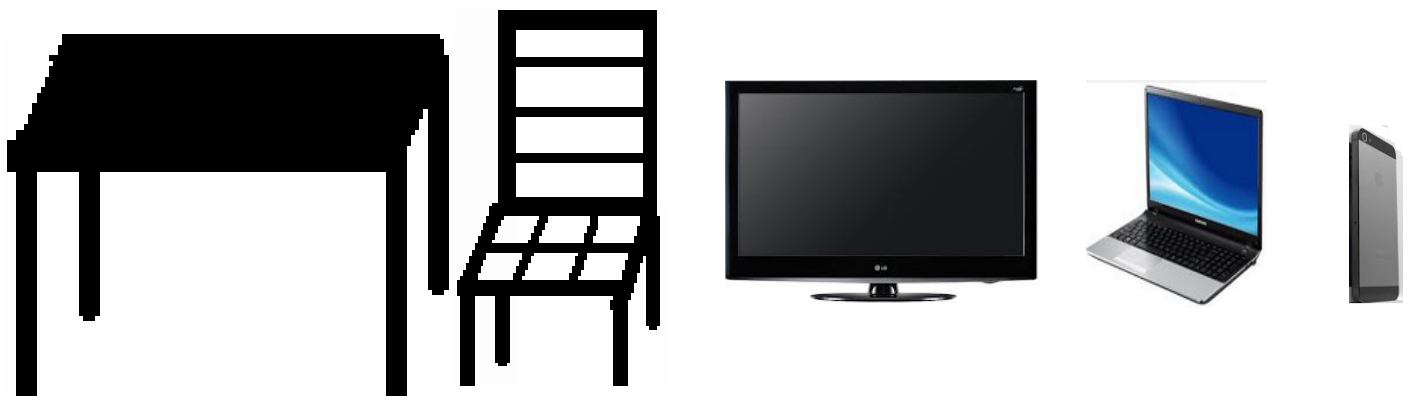

Figure3. Cluster1: objects having shapes of a 'rectangle' (Table, Chair, TV, Laptop, Mobile)

## 4.2. Detect based on the Scale of an object

Different objects have different scales to represent them. For example, the approximate scale of a Cup could be 16x16 while a ball could be 4x4 pixels. General studies have taken similar scale for all different objects. In most of the cases, by taking larger scale for smaller objects increases the computation timing during any image processing process either in several phases of training or testing. However knowing the most approximate scale of a query image is required. During training and testing phases the live edge detection and background segmentation can be performed.

The technique we used is a clustered window mapping. It has five different sub-scales windows. This technique will mapped the query image to a most appropriate scale. To avoid exhaustive search (comparison), the binary search techniques are incorporated. The samples of the five different sub-scaled windows are shown in figure4. Window 1 is the basic building block of other remaining windows. Generations of these windows are the computation result of either addition or multiplication of the window1 and/or transformation operation etc, with respect to the x-y plane. In needs, it can generate larger windows to map a larger object. To measure the limit of the larger object the boundary structure segmentation can be used. For our study, we have used 5 templates of windows for both training and testing processes. The first mappings are done with the window5, and then choose the most appropriate one by considering the sub-scales windows that maps the objects. Our study is mainly concentrated to detect the in-house objects; we preferred to have minimum to minimum spaces.

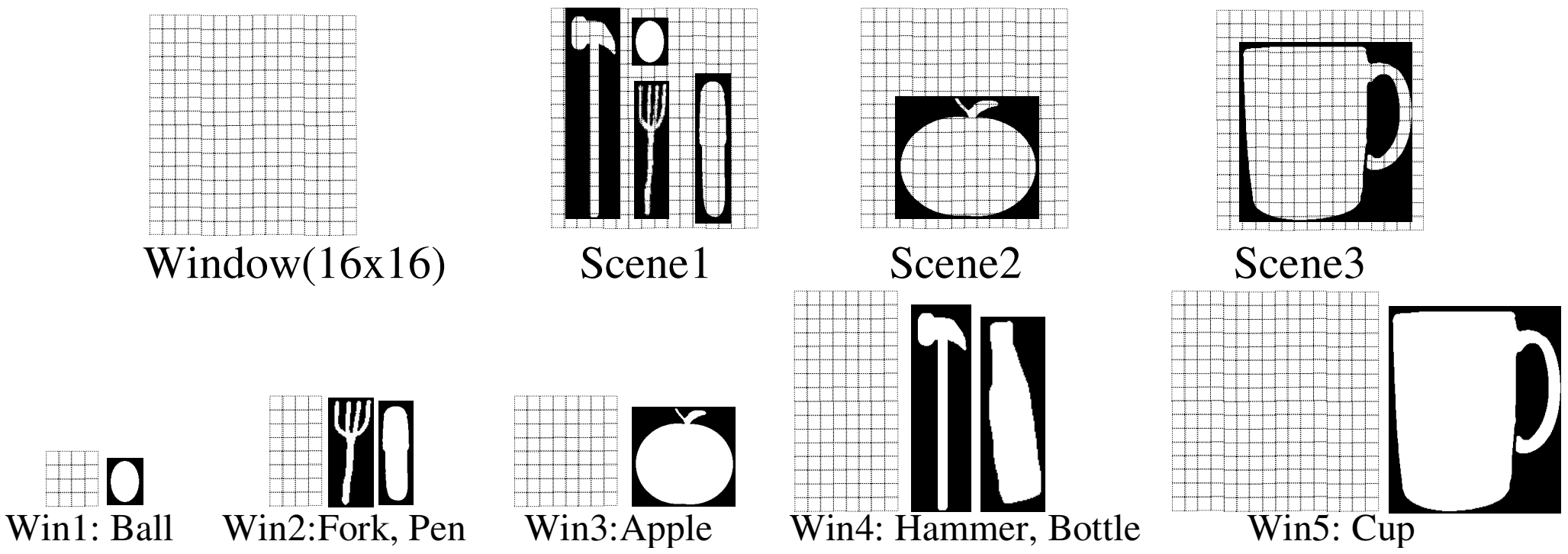

Figure4. Window1:[4x4], 2:[8x4], 3:[8x8], 4:[16x8], 5:[16x16], object clusters based on the scale(windows)

### 4.3. Detect object using the mean clustered values based on shapes

Modular approach is used to detect the object. A number of classifiers are used and each classifier is found suitable to classify a particular kind of feature vectors which depends upon their shape. Modular approaches partitions the classification task into some sub-classification and sub-sub classification, solve each sub-sub classification task and sub-sub classification and eventually integrates the result to obtain the final classification result. Clustering of items is done based on shape followed by the scale (size). Each of these clusters (having many items) has a mean value. The feature of a query image is first comparing with the mean value of clusters. The formula is given in equation1.

$$Cmv = \int_{1}^{n} \sum_{i=1}^{m} s.... \qquad ...(1)$$

Where
Cmv= mean value of a cluster.
n= No of total Clusters.
m= No of objects in a cluster those have a common shape.
s=different shapes (templates)

Some of the earlier techniques performs comparison operation generally based on mean features values [19, 20]. This study incorporated the shape factor also. Because each object has a different shape and scale, there could be several closed mean value of clusters which may have several items inside. This study gives faster computation time than other technique those use only mean values.

### 4.4. The proposed algorithm for feature selection (Training)

1. Input a scene
2. Remove noise
3. Segment the objects as per the shape, followed by scale
4. Select the features based on shape and followed by scale
5. Create a cluster for each different define shapes (define in our study)
6. Create a sub cluster within the cluster of shape for each different scale (define in our study)
7. Create an index1 to track the location of cluster based on shape, followed by scale

8. Load to database as per the shape and scale
9. Update the index1 for each object loaded to each cluster
10. Stop

The captured image may have noise, firstly noise is removed from the image whether i.e. salt & pepper noise, white & black noise etc. Then the image is segmented into number of objects present in the image, these segmented objects are classified into clusters as per the shape which is followed by scale. Create a new cluster if it is not already present.

## 4.5. The proposed algorithm for Testing

1. Input a scene
2. Remove noise
3. Segment the objects as per the shape, followed by scale
4. Select the features based on 'shape' and followed by 'scale'
5. Find the cluster that contains the 'shape' of the query image using the Index1 created during training phase.
6. Find the sub cluster for possible scale(window size) of the query image from the Index1 created during training phase
7. Compare the feature of the query image to mean value of the clusters that are stored based on scale (size).
8. Detect the exact object
9. Stop

## 4.6. Scale space extrema detection using Gaussian rule

The algorithm of Lowe[18] is used to detect and describe local features in images. The features are invariant to scale, translation and rotation which can achieve keypoints of image. By using the keypoints, objects in a scene can recognize and identify in the other images. It extracts the keypoints that are invariant to changes of scale.

Function, $L(x, y,\sigma)$, is the scale space(s) of an object(s) that obtained from the convolution of an input image, $I(x,y)$, with a variable scale-space Gaussian function, $G(x,y,\sigma)$. To efficiently detect stable keypoint locations, Lowe has proposed using the scale-space extrema in difference-of-Gaussian (DoG) function, $D(x,y,\sigma)$. Extrema computed by the difference of two nearby scales separated by a constant factor K (1).

$$\begin{aligned} D(x,y,\sigma) &= (G(x,y,k\sigma) - (G(x,y,\sigma))*I(x,y) \\ &= L(x,y,k\sigma) - L(x,y,\sigma) \qquad \ldots..(2) \end{aligned}$$

This process will be repeated in several octaves. In each octave, the initial image is repeatedly convolved with the Gaussian function to produce the set of scale space images. The adjacent Gaussian images are subtracted to get the difference of gaussian image(s). After each octave, the Gaussian images are down-sampled by a factor of 2, and the process is repeated. The combination of this algorithm with the mean clustering and or the clustering based on size also enhances the performance of the system.

## 4.7 Computation Time based on shape

A sample of performance of study is given as figure5. It shows the processing time taken during an exhausted search and search time using the 'shaped based' approach. Generally the time taken during the non-shaped based (exhausted search) took thrice the original time. By introducing such study on the object detection based on the shaped give speedy processing time. It can save time during testing process any (or more) objects.

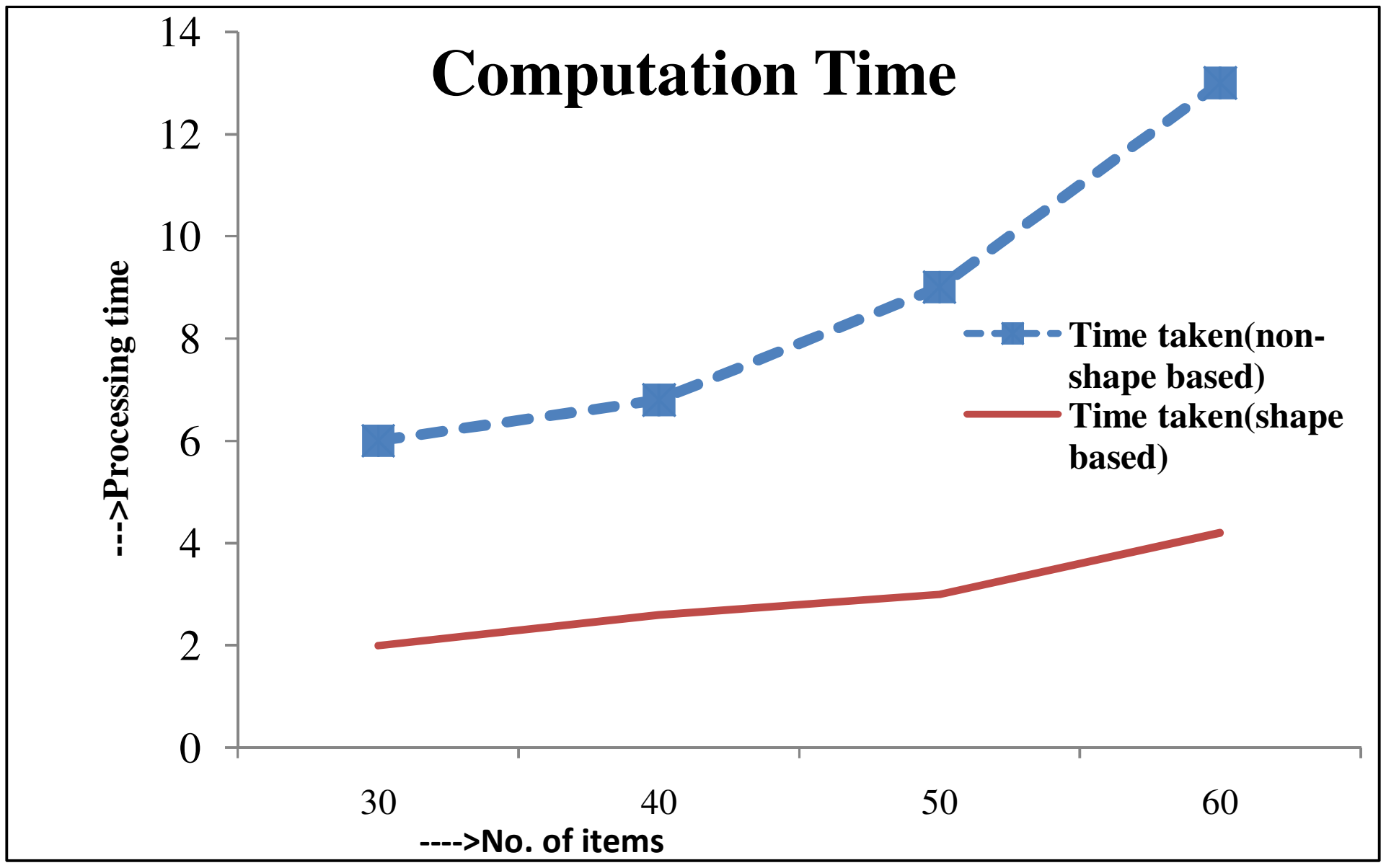

Figure5. Computation time

These are only the statistical data. The study is an ongoing study and many more improved results are expecting.

## 5. CONCLUSIONS

An in-house object detection system is designed. The object detection uses the dynamic clustering and scaling of training images and testing images. Algorithms of training and testing process are proposed. The shape and scale of the query image is not matching with shapes of objects in the available database (training), the remaining several operation of object detection: segmentation, cleaning, normalization, detection etc will not perform. This will save computation time. Creating the dynamic clusters based on the shape of the object and also sub cluster based on the scale (size) of the objects help to have a speedy feature selection and object detection. The concept of indexing the location of shape in each cluster and concept of mean value of clusters incorporated during searching of the features or database objects give faster speed.

The combine effect of execution of the algorithm of Lowe (which was developed to detect and describe local features in combining images) with the mean clustering and or the clustering based on size also enhances the performance of the system. Earlier techniques were generally based on mean features values comparisons. This study detect object using the mean clustered values based on shapes incorporating the shape factor gives speedy detection of objects. There could be several closed mean value of clusters which may have several items inside. The study finds different ways of speedy detection of objects by combination of clustering based on shape, size and means values.

## ACKNOWLEDGEMENTS

This survey report is a part of AICTE sponsor project. We deeply acknowledged AICTE (Govt. of India) for sponsoring this research work.

## REFERENCES

[1] C. Capelle, C. Trullemans, P. Arno, and C. Veraart, A Real-Time Experimental Prototype for Enhancement of Vision Rehabilitation Using Auditory Substitution, IEEE Trans. Biomed. Eng., Vol. 45, pp. 1279-1293, Oct. 1998.
[2] Shoval S. Auditory guidance with the Navbelt: a computerized travel aid for the blind. IEEE transactions on SMC, Part C, , 28(3): 459-467, 1998
[3] Ram S, etal. The PeopleSensor: A Mobility Aid for theVisually Impaired. IEEE Second International Symposium Wearable Computers, 166-167, 1998
[4] Ulrich I, Borenstein J. The GuideCane – applying mobile robot technologies to assist the visually impaired. IEEE Transactions on Systems, Man and Cybernetics, Part A, 31(2):131-136, 2001
[5] Bourbakis NG, Kavraki D. An intelligent assistant for navigation of visually impaired people, Proceedings of the IEEE 2nd International Symposium on Bioinformatics and Bioengineering Conference. 230-235, 2001
[6] Kuc R, Binaural Sonar Electronic Travel Aid Provides Vibrotactile Cues for Landmark, Reflector Motion and Surface Texture Classification. IEEE Transactions on Biomedical Engineering, 49(10): 1173-1180, 2002,
[7] Meijer P, An experimental system for auditory image representations. IEEE Transactions on Biomedical Engineering, 112-121, 1992,
[8] Milios E, Kapralos B, Kopinska A, Stergiopoulos S. Sonification of range information for 3-D space perception. IEEE Transactions on Neural Systems and Rehabilitation Engineering, 11(4): 416 – 42, 2003
[9] K Sawa, K Magatani, K Yanashima. Development of a navigation system for visually impaired persons by using optical beacons. Proceedings of second joint EMBS/BMES conference, 2426-2427, 2002
[10] B Ando. Electronic Sensory Systems for the Visually Impaired. IEEE Instrumentation & Measurement Magazine.6(2):62-67, 2003
[11] 11.G. Sainarayanan, R. Nagarajan, S. Yaacob. Fuzzy image processing scheme for autonomous navigation of human blind. Applied Soft Computing, 7(1): 257-264, Jan. 2007
[12] XU Jie, FANG Zhigang. Aud ioM an: Design and Implementation of Electronic Travel Aid. Journal of Image and Graphics, China. July, 12(7): 1249-1253, 2007
[13] Min Nie, Jie Ren, Zhengjun Li. SoundView: An Auditory Guidance System Based on Environment Understanding for the Visually Impaired People. Conference of the IEEE, EMBS, 7240-7243, 2009
[14] Jin Liu, Jingbo Liu, Luqiang Xu, Weidong Jin, Electronic travel aids for the blind based on sensory substitution. Computer science and education, pp 1328–1331, 2010
[15] Biederman, ”Recognition-by-components: A theory of human image understanding”. Psychological Review, 94(2):115–147, 1987
[16] Ashraf Elnagara, Reda Alhajjb, Segmentation of connected handwritten numeral strings, Pattern Recognition 36, 625 – 634, 2003
[17] LLE00- L.J. Latecki, R. Lakamper, and U. Eckhardt. Shape descriptors for non-rigid shapes with a single closed contour. IEEE Conf. on Computer Vision and Pattern Recognition, 5:424 429, 2000
[18] David G. Lowe, ”Object Recognition from Scale Invariant Features”, International Conference of Computer Vision, PP: 1150-1157, 1999
[19] V. Ferrari, F. Jurie, and C. Schmid, From Images to Shape Models for Object Detection International Journal of Computer Vision,2009
[20] Grimaldo Silva, Leizer Schnitman, Luciano Oliveira, "Multi-Scale Spectral Residual Analysis to Speed up Image Object Detection," sibgrapi, pp.79-86, 2012 25th SIBGRAPI Conference on Graphics, Patterns and Images, 2012

## Authors

**Y. Jayanta Singh** is working as Associate Professor and Head of Department of Computer Science & Engineering and Information Technology, Don Bosco College of Engineering and Technology of Assam Don Bosco University. He has received his Ph.D from Dr.B.A Marathwada University in 2004. He has worked with Swinburne University of Technology (AUS) at Malaysia campus, Misurata University, Keane (India and Canada) etc. His expertise area is Real time Database system, Image processing and Software Engineering practices. He has produced several papers in International and National Journal and Conferences.

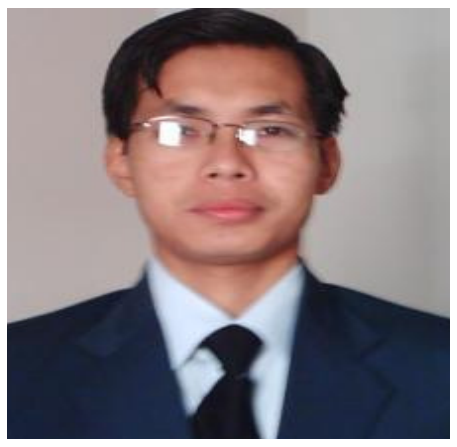

**Shalu Gupta** received her MCA and M. Tech degree in Computer Science from Maharshi Dayanad University, Rohtak and Lovely Professional University, Phagwara in 2007 and 2010 respectively. Presently working as a research scholar at Assam Don Bosco University and her areas of interest include image processing and pattern recognition.

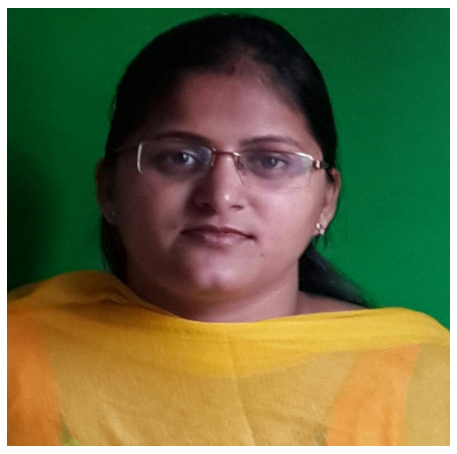